\documentclass[runningheads]{llncs}
\usepackage[T1]{fontenc}
\usepackage{graphicx}
\usepackage{amsfonts}
\usepackage{bbm}
\usepackage{pgfplots} 
\usepackage{adjustbox}
\usepackage{comment}
\usepackage[caption=false]{subfig} 
\usepackage{nicematrix}
\usepackage{amsopn}
\usepackage{booktabs} 
\usepackage{amsmath}
\usepackage{algorithm}
\usepackage{algpseudocode}

\DeclareMathOperator*{\argmin}{argmin}

\usepackage{hyperref}
\hypersetup{
    colorlinks=true,
    linkcolor=blue,
    citecolor=green,
    filecolor=black,
    urlcolor=black,
}

\begin{document}
\title{A model is worth tens of thousands of examples
}
%
%
\author{Thomas Dag\`es\inst{1}
\and
Laurent D. Cohen\inst{2} \and
Alfred M. Bruckstein\inst{1}}
%
\authorrunning{T. Dag\`es et al.}
%
\institute{Department of Computer Science, Technion Israel Institute of Technology, Haifa, Israel\\ \email{\{thomas.dages, freddy\}@cs.technion.ac.il}
\and
Ceremade, University Paris Dauphine, PSL Research University, UMR CNRS 7534, 75016 Paris, France\\ \email{cohen@ceremade.dauphine.fr}}
%
\maketitle              
\begin{abstract}

Traditional signal processing methods relying on mathematical data generation models have been cast aside in favour of deep neural networks, which require vast amounts of data. Since the theoretical sample complexity is nearly impossible to evaluate, these amounts of examples are usually estimated with crude rules of thumb. However, these rules only suggest when the networks should work, but do not relate to the traditional methods. In particular, an interesting question is: how much data is required for neural networks to be on par or outperform, if possible, the traditional model-based methods? In this work, we empirically investigate this question in two simple examples, where the data is generated according to precisely defined mathematical models, and where well-understood optimal or state-of-the-art mathematical data-agnostic solutions are known. A first problem is deconvolving one-dimensional Gaussian signals and a second one is estimating a circle's radius and location in random grayscale images of disks. By training various networks, either naive custom designed or well-established ones, with various amounts of training data, we find that networks require tens of thousands of examples in comparison to the traditional methods, whether the networks are trained from scratch or even with transfer-learning or finetuning.

\end{abstract}

\section{Introduction} 

Neural network-based machine learning has widely replaced the traditional methods for solving many signal and image processing tasks that relied on mathematical models for the data \cite{lecun2015deep,goodfellow2016deep}. In some cases, the assumed models provided ways to optimally address the tasks at hand and resulted in well-performing estimation and prediction methods with theoretical guarantees \cite{wiener1949extrapolation,novikoff1963convergence,elad2010sparse}. Nowadays, gathering raw data and applying gradient descent-like processes to neural network structures \cite{lecun1989handwritten,krizhevsky2017imagenet,simonyan2014very,he2016deep} largely replaced modelling and mathematically developing provably optimal solutions. 

It is commonly accepted that, if the networks are complex enough and when vast amounts of data are available, neural networks outperform traditionally designed methods \cite{mohamed2009deep,krizhevsky2017imagenet} or even humans \cite{geirhos2018generalisation,geirhos2021partial,bengio2021deep}. The required amount of data is called in statistical learning theory the sample complexity and is related to the VC-dimension of the problem \cite{vapnik2015uniform}, which is usually intractable for non trivial networks \cite{anthony_bartlett_1999}. Instead, various rules of thumb have been used in the field to guess how many samples are needed: at least 10-50 times the number of parameters \cite{alwosheel2018your}, at least 10 times per class in classification (and 50 times in regression) the data dimensionality \cite{lakshmanan2020machine} and at least 50-1000 times the output dimension \cite{alwosheel2018your}.

However, these rules only suggest how much data is needed to get a ``good'' network, but they do not relate to the traditional data-generation model-based methods. A natural question hence arises: do the neural network-based solutions perform as well as, or even outperform, the processing methods based on traditional data-generation models when lots of data is available, and if so how much data is necessary? We address this question in two simple empirical examples, where the data is produced according to precisely defined models, and where well understood optimal or state-of-the-art mathematical solutions are available. The first is the deconvolution of Gaussian signals, optimally solved with the Wiener filter \cite{wiener1949extrapolation}. The second is the estimation of the radius and centre coordinates of a disk in an image, which can be elegantly solved using a Pointflow method \cite{yang2017model}. This work aids engineers to decide when to use model-based classical methods or simply feed lots of data (if available) to deep neural networks.

Section \ref{sec: 1D} presents our comparison for the one-dimensional signal recovery, and Section \ref{sec: 2D} deals with estimation of disk characteristics in an image.

\section{One-dimensional signal recovery}

\label{sec: 1D}

We suggest to first analyse a simple and well-understood problem in the one-dimensional case where the optimal solution is provingly known.

\subsection{Data model and optimal solution}

The original data consists of real random vectors $\varphi$ of size $D$ that are centred, i.e. $\mathbb{E}(\varphi)=0$, and with known autocorrelation $R_\varphi = \mathbb{E}(\varphi\varphi^\top)\in\mathbb{R}^{D\times D}$. However, $\varphi$ is degraded by blur and noise producing the observed data $\varphi^{\mathit{data}}$ as follows:
\begin{equation}
    \label{eq: 1d data model}
    \varphi^{\mathit{data}} = H\varphi + n,
\end{equation}
where $H\in\mathbb{R}^{D\times D}$ is a known deterministic matrix and $n$ is random additive noise independent from $\varphi$ that is centred $\mathbb{E}(n) = 0$ and with known autocorrelation matrix $R_n\in\mathbb{R}^{D\times D}$. It is well-known \cite{wiener1949extrapolation} that the best linear recovery of $\varphi$ in the $L^2$ sense, i.e. minimising the Expected Squared Error (ESE) $\mathit{ESE}(\hat{\varphi}, \varphi) = \mathbb{E}\left(\lVert \hat{\varphi} - \varphi \rVert_2^2\right)$ with respect to the matrix $M\in\mathbb{R}^{D\times D}$ such that $\hat{\varphi} = M\varphi^{\mathit{data}}$, is given by applying the Wiener filter $W = R_\varphi H^\top(H R_\varphi H^\top + R_n)^{-1}$, i.e. $\hat{\varphi}^* = W\varphi^{\mathit{data}}$. Moreover, if we further assume both $\varphi\sim\mathcal{N}(0,R_\varphi)$ and $n\sim \mathcal{N}(0,R_n)$ are Gaussian, then the Wiener filter $W$ minimises the ESE over all possible recoveries including nonlinear ones. Furthermore, note that if $R_\varphi$ is circulant, i.e. $\varphi$ is cyclostationary, and so is $n$, e.g. if $n$ has independent entries implying $R_n$ is diagonal, and if $H$ is circulant, then $W$ is also circulant and Wiener filtering is a pointwise multiplication in the Fourier domain given by applying the unitary Discrete Fourier Transform $[DFT]$ with $(k,l)$-th entry $[DFT]_{k,l} = \frac{1}{\sqrt{D}} e^{-i\frac{2\pi kl}{D}}$.

\begin{figure}[!ht]
    \vspace{-1em}  
    \centering
    \subfloat{
         \includegraphics[width=0.28\textwidth]{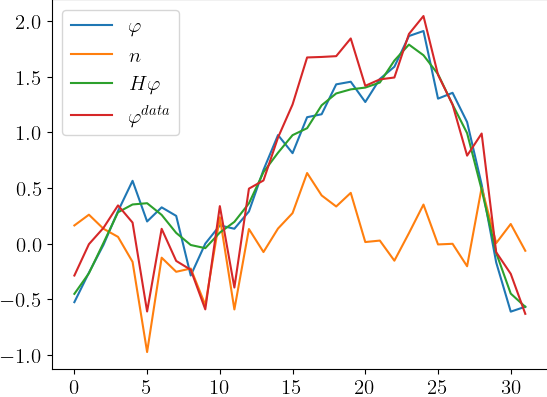}}
    \quad\quad\quad\quad
    \subfloat{
         \includegraphics[width=0.28\textwidth]{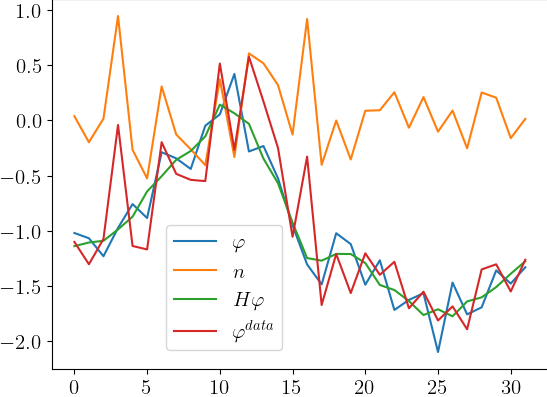}}
\caption{Two example signals $\varphi^{data}$, with their associated blur $H\varphi$ and noise $n$.}
\label{fig: 1d data}
\end{figure}

\begin{figure}[!ht]
    \vspace{-2em}  
    \centering
    \captionsetup[subfigure]{labelformat=empty}
    \subfloat[$H$]{
         \includegraphics[width=0.12\textwidth]{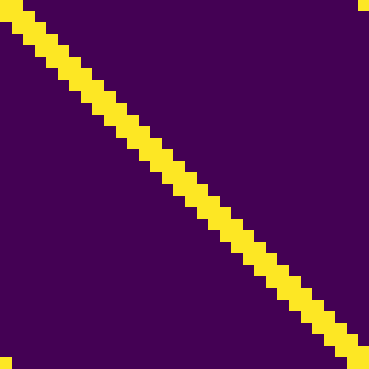}}
    \quad\quad
    \subfloat[$R_\varphi$]{
         \includegraphics[width=0.12\textwidth]{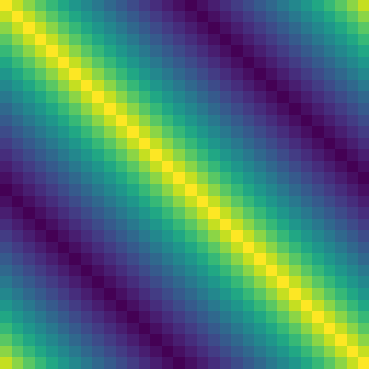}}
    \quad\quad
    \subfloat[$R_n$]{
         \includegraphics[width=0.12\textwidth]{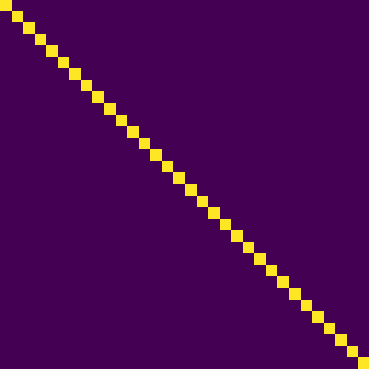}}
    \quad\quad
    \subfloat[$W$]{
         \includegraphics[width=0.12\textwidth]{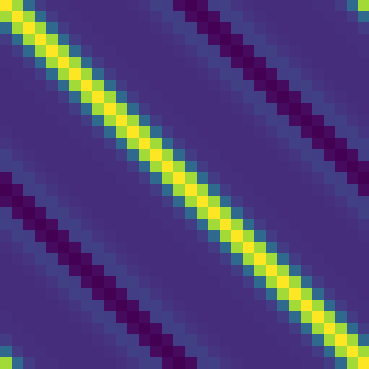}}
    \hfill\\
\caption{Chosen model matrices and associated optimal Wiener filter.}
\label{fig: 1d H R_phi R_n W}
\end{figure}

In our tests, the dimensionality is $D=32$ and the problem is circulant. We use an interpretable symmetric positive-definite autocorrelation matrix $R_\varphi$ parameterised by a large number $\rho = 0.95$ to create high spatial correlation over a large support decaying with distance and $H$ is a local smoothing convolution.
The first lines of $R_\varphi$ and $H$ are
$\begin{pNiceMatrix} 1&\rho&\rho^2&\rho^3&\Cdots&\rho^3&\rho^2&\rho\end{pNiceMatrix}$
and $\begin{pNiceMatrix}1 & 1 & 0 &\Cdots[line-style=solid] & 0 &1 \end{pNiceMatrix}$.
The noise is i.i.d. $n\sim\mathcal{N}(0,\sigma_n^2 I)$ with $\sigma_n =0.1$.
We display example data in Figure \ref{fig: 1d data}, the designed $H$, $R_\varphi$, and $R_n$ along with their associated Wiener filter $W$ in Figure \ref{fig: 1d H R_phi R_n W}.

\subsection{Neural models}

We wish to evaluate the capabilities of neural networks by comparing them to humanly designed methods by classical experts using no training. Our criterion is the amount of random training samples $N$ needed to reach or overtake human expertise. Working in the Gaussian case for the data model of equation (\ref{eq: 1d data model}), we create various random training datasets containing $N$ data samples ranging in $N\in\{10,100,1000,10000,100000\}$. We train a variety of small Convolutional Neural Networks (CNNs) of various depths $K\in\{0,1,2,3\}$. The depth of the network is measured as the number of successions of convolution-pointwise-nonlinearity layers. Each network ends with a final fully connected layer $A$ (with bias $b_A$), i.e. a final unconstrained affine transformation. For simplicity, our CNNs will be single-channel only and without various architecture tricks, e.g. dropout, batch normalisation, or pooling.  The network functions, denoted $f_k$ for $k\in\{1,\ldots,K\}$ can thus be written as:
\begin{equation}
    f_k(\varphi^{\mathit{data}}) = A \sigma\circ  \tilde{C}_K \circ \sigma\circ \tilde{C}_{K-1} \circ \cdots \circ \sigma\circ \tilde{C}_1 (\varphi^{\mathit{data}})+ b_A,
\end{equation}
where $\tilde{C}_i(x) = C_i x + b_i$ is the $i$-th convolution layer comprising the circulant matrix $C_i$ for the convolution and its additive unconstrained bias $b_i$ and $\sigma=\mathrm{ReLU}$ the standard pointwise nonlinearity in neural networks. Note that a CNN with depth $0$ degenerates to an unconstrained affine transformation in $\mathbb{R}^D$ (no pointwise nonlinearity or convolution): $f_0(\varphi^{\mathit{data}}) = A\varphi^{\mathit{data}} + b_A$.


The networks are trained to minimise the Mean Squared Error (MSE)\footnote{The loss function is actually scaled to $\tfrac{1}{D}\mathit{MSE}_{\mathit{train}}$ as is commonly done in practice.}, a proxy for the ESE, using the $N$ generated samples. Denoting $f_{k,N,\eta}$ the resulting networks (where $\eta$ a hyperparameter of the optimisation algorithm), we have:
\begin{equation}
    \mathit{MSE}_{\mathit{train}}(f_{k,N,\eta}) = \tfrac{1}{N}\sum\limits_{i=1}^N \lVert f_{k,N,\eta}(\varphi_{\mathit{train},i}^{\mathit{data}}) - \varphi_{\mathit{train},i} \rVert_2^2,
\end{equation}
where for a sample collection $\mathit{set}$, $\varphi_{\mathit{set},i}$ and $\varphi_{\mathit{set}, i}^{\mathit{data}}$ denote the $i$-th original and degraded samples. This quantity is to be compared with $\mathit{ESE}(f_{k,N}(\varphi^{\mathit{data}}),\varphi)$, which evaluates the performance on all possible data of a network trained on $N$ instances only. Naturally, this quantity cannot be computed by hand and is approximated by another $\mathit{MSE}$ calculation on a large test set using $N_{t}$ test samples independently generated from the training ones:
\begin{equation}
    \mathit{MSE}_{\mathit{test}}(f_{k,N,\eta}) \! = \! \tfrac{1}{N_{t}}\!\sum\limits_{i=1}^{N_t} \lVert f_{k,N,\eta}(\varphi_{\mathit{test},i}^{\mathit{data}}) - \varphi_{\mathit{test},i} \rVert_2^2 \! \xrightarrow[N_t\to\infty]{} \! \mathit{ESE}(f_{k,N,\eta}(\varphi^{\mathit{data}}),\varphi).
\end{equation}
In our tests, $N_{t} = 100000$. Note that implicitly in $\mathit{ESE}(f_{k,N,\eta}(\varphi^{\mathit{data}}),\varphi)$ the network $f_{k,N,\eta}$ is the given result of a minimisation algorithm. For randomised algorithms, it is thus to be understood as the expectation conditional to the learned network $f_{k,N,\eta}$: $ \mathit{ESE}(f_{k,N,\eta}(\varphi^{\mathit{data}}),\varphi) =  \mathbb{E}(\lVert f_{k,N,\eta}(\varphi^{\mathit{data}}) -\varphi\rVert_2^2 \mid f_{k,N,\eta})$.

We train our networks using Stochastic Gradient Descent with Nesterov momentum parameter equal to $0.9$. 
We train the networks using various learning rates $\eta\in\{0.0001, 0.0005, 0.001, 0.005, 0.01, 0.05, 0.1\}$
over $50$ epochs, performing $N_r = 50$ independent training trials per learning rate, and compute the final median performance per learning rate on a validation set generated independently of the train and test data comprising $N_{v} = 100000$ validation samples:
\begin{equation}
    \mathit{MSE}_{\mathit{val}}(f_{k,N,\eta}) \!=\! \tfrac{1}{N_{v}}\sum\limits_{i=1}^{N_v} \lVert f_{k,N,\eta}(\varphi_{\mathit{val},i}^{\mathit{data}}) - \varphi_{\mathit{val},i} \rVert_2^2\xrightarrow[N_v\to\infty]{} \mathit{ESE}(f_{k,N,\eta}(\varphi^{\mathit{data}}),\varphi).
\end{equation}
The validation set is used to choose the best learning rate for each amount of training data $\eta^*(N)$ by taking:
\begin{equation}
    \eta^*(N) = \argmin\limits_{\eta} \mathit{MEDIAN}_r(\mathit{MSE}_{\mathit{val}}(f_{k,N,\eta})),
\end{equation}
where $\mathit{MEDIAN}_r$ takes the median over the $r\le N_r$ best independent runs on the validation set per $\eta$. Given that a significant amount of runs do not converge or get trapped early in a poor local minimum depending on the random initialisation, choosing $r\ll N_r$ ensures that only the networks finding a good local minimum are considered. The final performance of CNNs $SCORE_{k,r}(N)$ for each amount of data $N$ is then the median of the test performance over those selected $r$ trials\footnote{Selected on the validation set.} of the final test score at the chosen learning rate $\eta^*(N)$:
\begin{equation}
    SCORE_{k,r}(N) = MEDIAN_r(\mathit{MSE}_{\mathit{test}}(f_{k,N,\eta^*(N)})).
\end{equation}

\begin{figure}[h]
    \vspace{-1em}  
    \centering
    \subfloat{\includegraphics[width=0.35\textwidth]{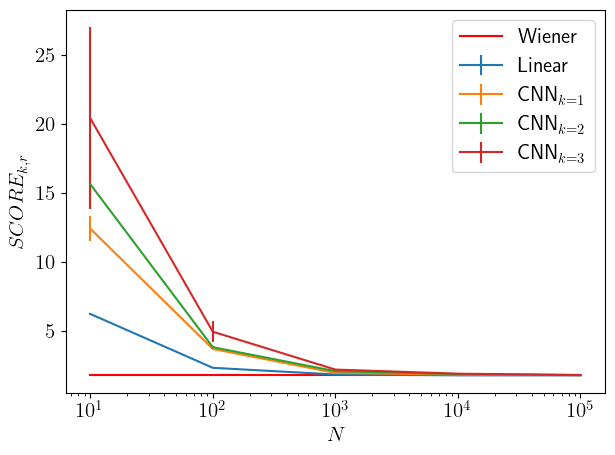}}%
    \quad\quad
    \subfloat{\includegraphics[width=0.35\textwidth]{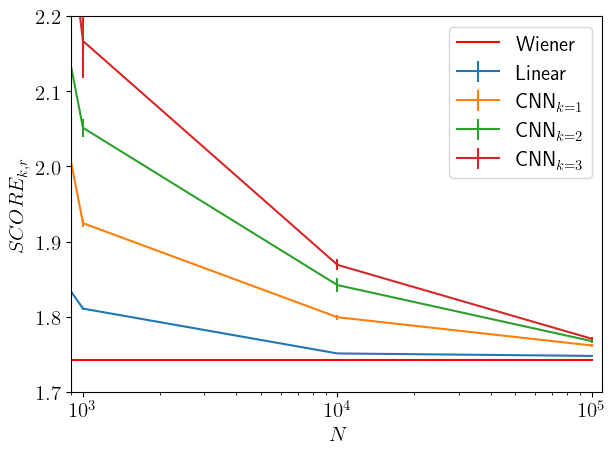}}%
\caption{Median test scores for CNNs with depth $k\in\{0,1,2,3\}$ on $r=10$ selected runs ($k=0$ is just a linear layer). Vertical bars represent the standard deviation of the MSE of these runs. The right figure is a zoomed-in plot of the left one for large $N$.}
\label{fig: 1d MSE_function_N_Bestk_10_median_test_on_best_median_val}
\end{figure}

We display the evolution of the networks' performance on the amount of training data $N$ in Figure \ref{fig: 1d MSE_function_N_Bestk_10_median_test_on_best_median_val} for each depth $k$, with detailed scores in Table \ref{tab: 1d median mse test r 10}, along with the performance of the Wiener filter. Regardless of $N$, the Wiener filter outperforms the neural models as expected by the theory, but their performance converges to the Wiener's one when a lot of data is available, with similar performance when at least $10000$ training samples are available. 
We can thus consider this study as providing a criterion that a model would be preferable if data is limited to fewer than 10000 samples to train on.

\begin{table}[ht]
    \centering
    \resizebox{0.5\textwidth}{!}{%
        \begin{tabular}{lcccccc}
            \toprule
                $N$ & 0 & 10 & 100 & 1000 & 10000 & 100000\\
                 \midrule
                Wiener         & \bf{1.743} & \bf{---} & \bf{---} & \bf{---} & \bf{---} & \bf{---} \\
                Linear ($k=0$) & ---& 6.204 & 2.295 & 1.811 & 1.751 & 1.748 \\
                CNN ($k=1$)    & ---& 12.386 & 3.662 & 1.924 & 1.799 & 1.762 \\
                CNN ($k=2$)    & ---& 15.614 &3.789 & 2.051 & 1.842 & 1.767 \\
                CNN ($k=3$)    & ---& 20.395 &4.911 & 2.167 & 1.869 & 1.771 \\
            \bottomrule
        \end{tabular}%
    } 
    \vspace{10pt}
    \caption{Median MSE scores $SCORE_{k,r}$ of the CNNs on $r=10$ selected runs, compared to the theoretically optimal Wiener filter.}
    \label{tab: 1d median mse test r 10}
\end{table}

\section{Two-dimensional geometric estimation}
\label{sec: 2D}

We next analyse a more complicated yet well-understood problem based on Euclidean geometry. The goal is to estimate basic geometric properties on simple data: the radius and centre location of a random disk in an image. 
It was shown in \cite{dages2023compass} that this seemingly trivial task is more complex than expected for neural models even when focusing on radius estimation of centred disks.

\subsection{Data model} 
\begin{figure}[!ht]
    \centering
    \subfloat{
         \includegraphics[width=0.11\textwidth]{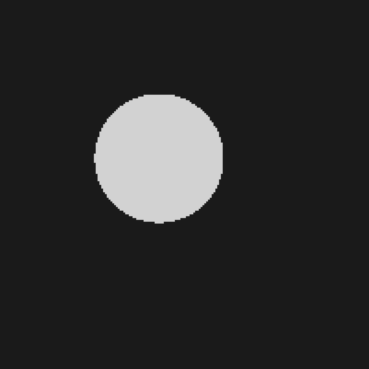}}
    \subfloat{
         \includegraphics[width=0.11\textwidth]{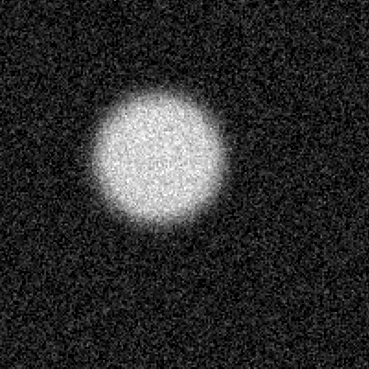}}
    \hfill
    \subfloat{
         \includegraphics[width=0.11\textwidth]{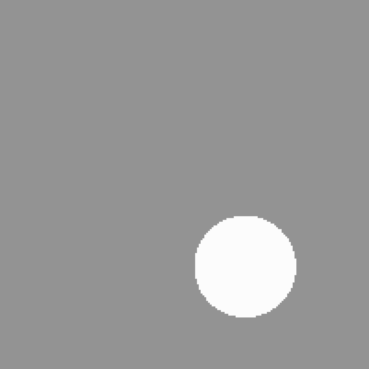}}
    \subfloat{
         \includegraphics[width=0.11\textwidth]{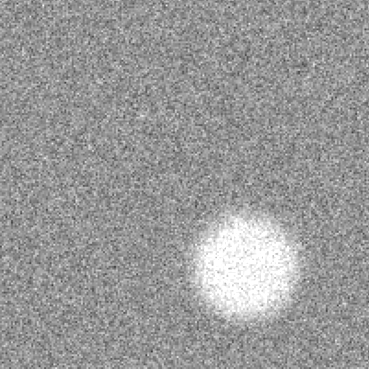}}
    \hfill
    \subfloat{
         \includegraphics[width=0.11\textwidth]{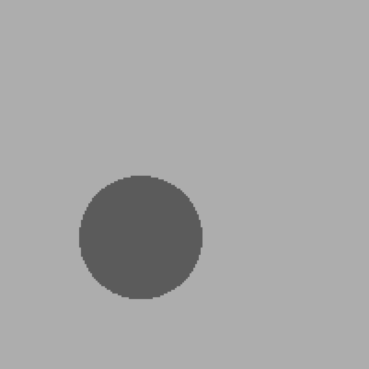}}
    \subfloat{
         \includegraphics[width=0.11\textwidth]{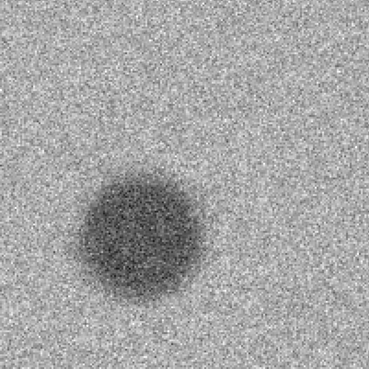}}
    \hfill
    \subfloat{
         \includegraphics[width=0.11\textwidth]{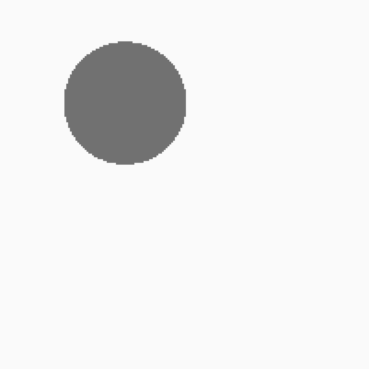}}
    \subfloat{
         \includegraphics[width=0.11\textwidth]{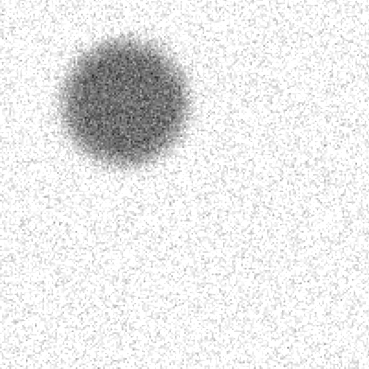}}
    \hfill
\caption{Four examples of clean $\varphi$ and degraded $\varphi^{data}$ disk images.}
\label{fig: circle data}
\end{figure}

The original data now consists of $D\times D$ random two-dimensional grayscale images of disks. Images are centred at $(0, 0)$, and for a pixel $x\in [-\tfrac{D-1}{2}, \tfrac{D-1}{2}]^2$:
\begin{equation}
    \varphi(x) = 
    \begin{cases}
        b & \text{if } \lVert x-c\rVert_2 > r\\
        f & \text{if } \lVert x-c\rVert_2 \le r,
    \end{cases}
\end{equation}
where $r$ is the circle's radius, $c = (c_x, c_y)$ its centre, and $f$ (resp. $b$) is the foreground (resp. background) intensity. These parameters are independently\footnote{Except $f$ and $b$ which are slightly correlated to ensure a minimal contrast $|f-b|>\delta$.} and uniformly chosen at random: $r\sim \mathcal{U}([\tfrac{\varepsilon_r}{2}\tfrac{D-1}{4}, (1-\tfrac{\varepsilon_r}{2})\tfrac{D-1}{4}])$ with $\varepsilon_r = 0.4$, $c\sim \mathcal{U}( [(D-1)\tfrac{\varepsilon_c}{2} - \tfrac{D-1}{2}, (D-1)(1-\tfrac{\varepsilon_c}{2}) - \tfrac{D-1}{2}]^2)$ with $\varepsilon_c = 0.5$, $b\sim \mathcal{U}([0,1])$, and $f\mid b\sim \mathcal{U}([0,1]\setminus [b-\delta, b+\delta])$ with  $\delta = \tfrac{50}{255}$ the minimum contrast\footnote{In our tests, we take $D=201$ implying that $r\sim\mathcal{U}([10,40])$ and $c\sim\mathcal{U}([-50, 50]^2)$.}. However, $\varphi$ is degraded with blur and noise giving the observed data $\varphi^{\mathit{data}}$ as follows:
\begin{equation}
    \label{eq: 2d data model}
    \varphi^{\mathit{data}} = g_{\sigma_b} * \varphi + n,
\end{equation}
where $g_{\sigma_b}(x) = \tfrac{1}{2\pi}\mathrm{exp}(-\tfrac{\lVert x\rVert_2^2}{2\sigma_b^2})$ is a Gaussian convolution kernel, and $n$ is i.i.d. white noise $n \sim \mathcal{N}(0, \sigma_n I_{D^2})$.
We plot example data in Figure \ref{fig: circle data}. The task is to estimate the three geometric numbers $(r,c) = (r, c_x, c_y)$ from $\varphi^{\mathit{data}}$.

\subsection{Expert engineer's solution} Unlike in the Wiener case, the optimal estimator minimising the ESE is not so trivial to find. Instead, we choose a method called Pointflow designed by an expert engineer that perfectly tackles the problem at hand. 

\begin{figure}[!ht]
    \centering
    \subfloat[$\varphi$]{\label{fig: circle pointflow J}
         \includegraphics[width=0.11\textwidth]{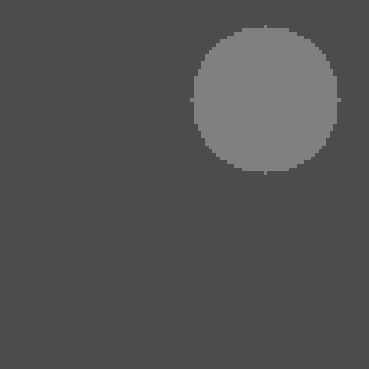}}
    \subfloat[$\varphi^{data}$]{\label{fig: circle pointflow I}
         \includegraphics[width=0.11\textwidth]{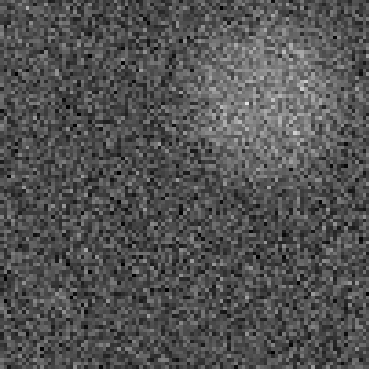}}
    \subfloat[$V_+$]{\label{fig: circle V+}
         \includegraphics[width=0.11\textwidth]{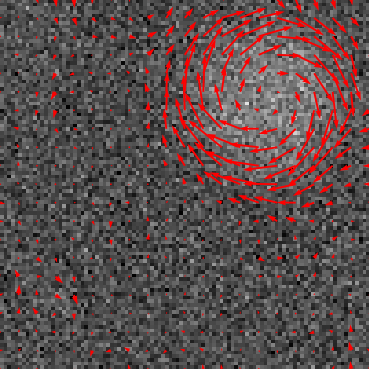}}
    \subfloat[$V_-$]{\label{fig: circle V-}
         \includegraphics[width=0.11\textwidth]{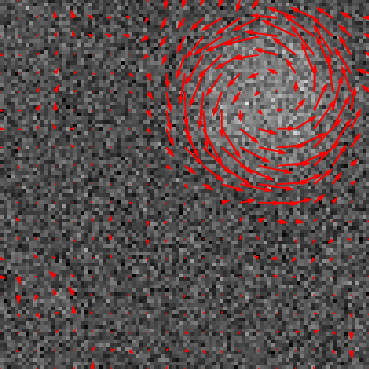}}
    \subfloat[]{\label{fig: circle no reflow}
         \includegraphics[width=0.11\textwidth]{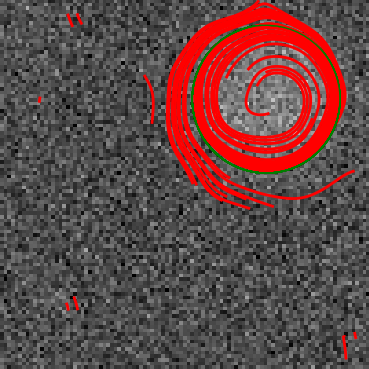}}
    \subfloat[]{\label{fig: circle only reflow e1}
         \includegraphics[width=0.11\textwidth]{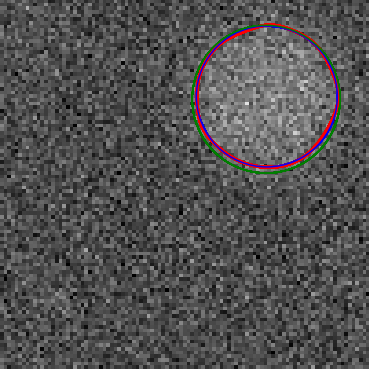}}
\caption{Pointflow in practice. (\ref{fig: circle pointflow J}): clean data. (\ref{fig: circle pointflow I}): degraded data. (\ref{fig: circle V+}) and (\ref{fig: circle V-}): pointflow fields sampled every five pixels. (\ref{fig: circle no reflow}): initial flows of points without reflowing with groundtruth boundary in green. (\ref{fig: circle only reflow e1}): all final trajectories that have reflown in a closed loop ($C_l$) with groundtruth boundary in green and all regressed circles using least squares on each looped trajectory in blue (used for estimating the centre).}
\label{fig: circle pointflow}
\end{figure}

Pointflow \cite{yang2017model,bai2019point} is an elegant subpixel level contour integrator and edge detector in images requiring no learning whatsoever. It consists in defining potential vector fields $V$ along which random points $P$  flow: $\tfrac{d P}{dt}(t) = V(P(t))$, such that end trajectories lie on edges of the image $I$. The vanilla Pointflow \cite{yang2017model} uses two fields $V_+$ and $V_-$ from the edge attraction $V_a$ and rotating $V_r$ fields based on the image gradients as follows:
\begin{equation}
    V_a = \nabla\lVert\nabla I_b\rVert_2, \quad
    V_r = \nabla I_b^\bot, \quad
    V_\pm = \tfrac{1}{2}(V_a \pm V_r),
\end{equation}
where $I_b = g_{\sigma_{Pf}}*I$ is a blurred version of $I$ with a Gaussian kernel $g_{\sigma_{Pf}}$. Various stopping conditions and uses of $V_\pm$ exist to detect edges in natural images, however on our data containing a single circular edge per image, we need only consider the basic ones. Indeed, the possible cases for trajectories are: it loops ($C_l$), it leaves the image domain ($C_o$), or it is stuck in an area with small magnitude $\lVert V\rVert_2$  ($C_s$). Flowing initially from $V_+$, if we loop ($C_l$), then the point has reached the circle and it suffices to reflow along $V_+$ to extract just the circle's contour. If we end up outside the image domain ($C_o$), which is rare in our data, we reflow from the exit point along $V_-$. If we are in a low flow magnitude area ($C_s$), which is not rare, then we discard the trajectory. In total, $N_{Pf} = 200$ points are randomly uniformly sampled in the image domain and used for Pointflow, and a fraction of them end up flowing on the disk's edge with subpixel precision, as the other ones lead to discarded trajectories. For more details on our implementation of Pointflow see Appendix \ref{sec: appendix pointflow details}. For some illustrations of Pointflow results on our data see Figure \ref{fig: circle pointflow}.

To estimate the disk's radius and centre from pointflow contours ($C_l$), we can simply compute the average length of the reflown closed contours and divide it by $2\pi$. To compute the centre's coordinates, we could compute for each closed trajectory the average of its points, and then average over these estimations. However, this method empirically did not best perform on validation data, so we refined it by applying least-squares regression on the equation of a circle to estimate from it its location per trajectory and then average the estimations. Note that the least-squares regression did not provide a better estimation of the radius so we keep the crude length integration strategy for it.

\subsection{Neural models} As in the one-dimensional case, the expert's method is to be compared with a convolutional neural network. Although the learning problem seems trivial, it is actually harder than expected for networks, as has been shown in \cite{dages2023compass} even when the circles are centred. Empirically, we were not able to train correctly a small custom model similar to those previously used having just three layers and even many channels per layers and no further deep learning tricks. To overcome this limitation, we use famous networks in the deep learning literature: Alexnet \cite{krizhevsky2017imagenet}, VGG \cite{simonyan2014very} and Resnet\footnote{
We use the simplest ones VGG11 and ResNet18, as larger ones are here unnecessary.
} \cite{he2016deep}. To adapt the model to our task, we change the final fully connected layer to have $3$ outputs only. 

For each architecture, we either train the networks from scratch (SC), or initialise the weights, except those of the final fully connected layers, to those publicly available obtained by classification on Imagenet \cite{russakovsky2015imagenet}, as is commonly done in the field. The pretrained weights can be either frozen for transfer-learning (TL) or retrained as well for finetuning (FT). Although the task and data are fundamentally different from ours, it is generally believed that the wide variety of natural images encourages the famous networks to learn features that generalise quite well to most reasonable tasks. 

Once again, the $\mathit{MSE}$ loss is used for training\footnote{To help the networks converge, the radius and centre coordinates are scaled to $[-1,1]$ using $r_s  = \tfrac{8}{D-1}(r-\tfrac{D-1}{8})$ and $c_s = \tfrac{2}{D-1}c$. In all plots and numbers provided in this paper, the results are rescaled to the original scale: $r$ and $c$ and not $r_s$ and $c_s$.}. As it is significantly more expensive to train such networks compared to the tiny ones in the Wiener case, we only perform $N_r = 1$ run per learning rate configuration, ranging in $\eta\in\{0.000001, 0.00005, 0.00001, 0.0005, 0.0001, 0.005, 0.001, 0.05, 0.01, 0.5, 0.1\}$, with a batch size of $10$ for $50$ epochs. As previously, the optimal learning rate is chosen on the performance on validation data. Both the test and validation data use $N_v = N_t = 100 000$ independently randomly sampled images, whereas the training sets have $N\in\{10,100,1000,10000,100000\}$ ones. 

\subsection{Results} 
\begin{figure}[!ht]
    \vspace{-1em}  
    \centering
    \captionsetup[subfigure]{labelformat=empty}
    \subfloat[$(r,c)$]{
         \includegraphics[width=0.29\textwidth]{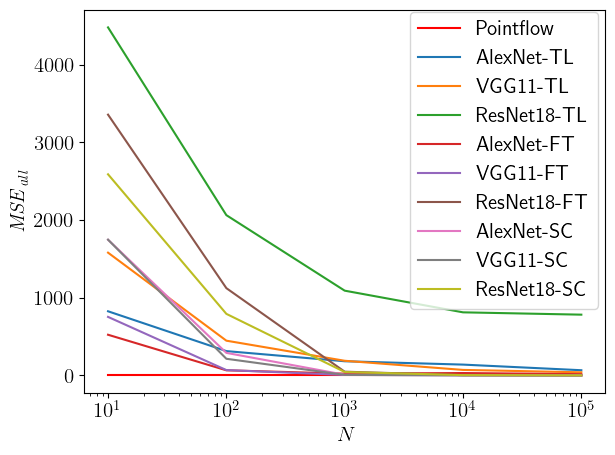}}
    \quad
    \subfloat[$r$]{
         \includegraphics[width=0.29\textwidth]{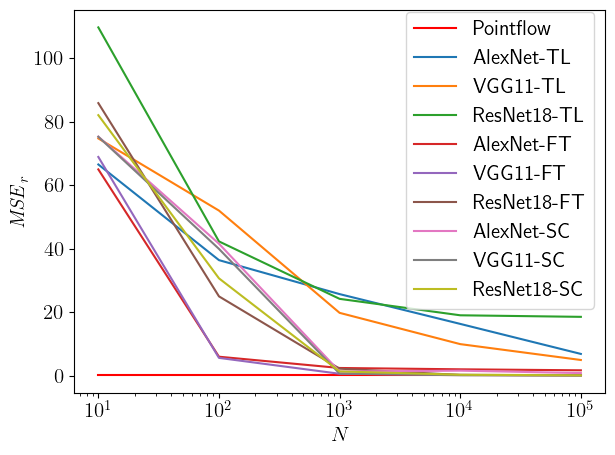}}
    \quad
    \subfloat[$c$]{
         \includegraphics[width=0.29\textwidth]{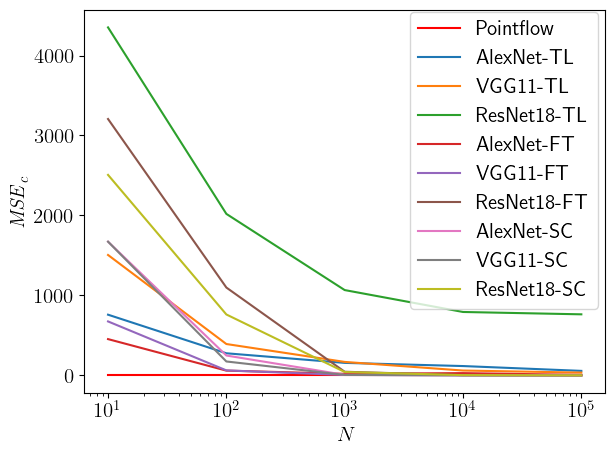}}
    \hfill\\
    \subfloat[$(r,c)$]{
         \includegraphics[width=0.29\textwidth]{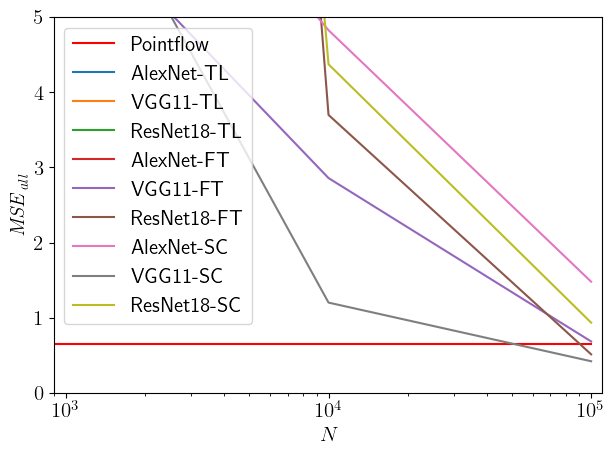}}
    \quad
    \subfloat[$r$]{
         \includegraphics[width=0.29\textwidth]{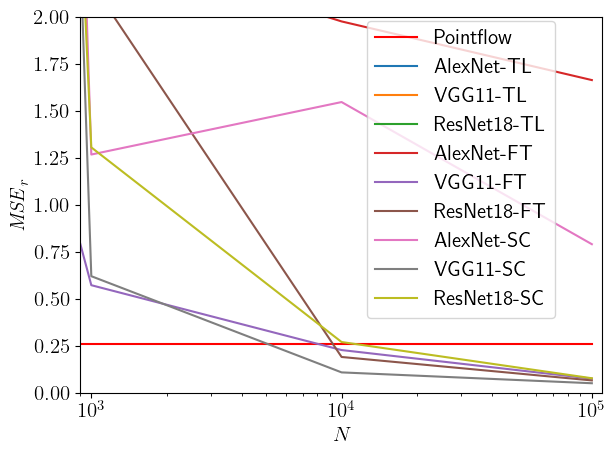}}
    \quad
    \subfloat[$c$]{
         \includegraphics[width=0.29\textwidth]{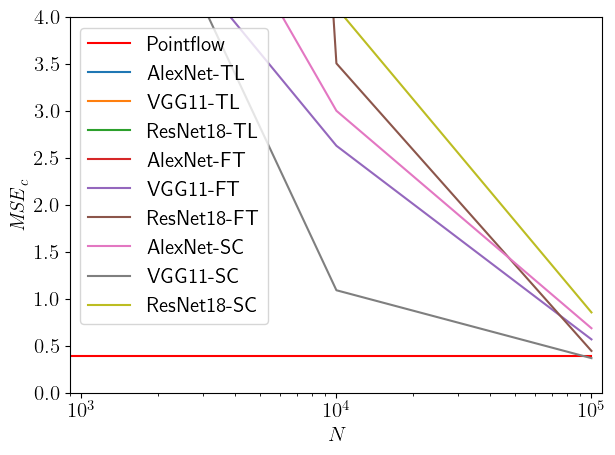}}
    \hfill\\
\caption{Test scores of the data-agnostic Pointflow and of the best transfer-learned or finetuned networks. Left: MSE computed on both the radius and the centre's location estimation. Middle: same but only on the radius estimation. Right: same but only on the centre location's estimation.  We zoom-in in the bottom set of figures. }
\label{fig: circles MSE_function_N_Bestk_1_median_test_on_best_median_val}
\end{figure}

We present the results in Figure \ref{fig: circles MSE_function_N_Bestk_1_median_test_on_best_median_val} and Table \ref{tab: circles mse test r 1}. Although the networks are simultaneously trained for both the radius and centre location estimation, we also present the MSE on the estimation of each geometric concepts separately.

First, the transfer-learning network Resnet-TL is not able to correctly estimate the radius or centre's location, meaning that its learned features on classification of Imagenet are not able to handle our simple data: they are not so general after all. Likewise, the prelearned features of Alexnet-TL and VGG-TL do not generalise well to this toy problem, requiring significantly more data than the maximum available to compare with the simple data agnostic pointflow. However, when the networks are entirely trained, either finetuned or from scratch, they are either flatly beaten by pointflow when using small amounts of training data or on par or slightly outperform it when $N\ge 10000$. The only networks beating pointflow overall are VGG-SC and Resnet-FT when $N=100000$, but more (VGG-FT, VGG-SC, Resnet-FT, and Resnet-SC) significantly outperform pointflow on the radius estimation when $N\ge 10000$. The difference between finetuning and training from scratch seems to only appear when small amounts of training data are available, and then finetuning is better. However, in these cases, both approaches pale in comparison to the reference data agnostic method.

From this experiment, we conclude that a realistic neural network is worth 
at least tens of thousands
of examples on a fairly simple task 
(toy problem
vs real world challenge)
compared to an expert engineer.
We thus provide a criterion that a data model is preferable if fewer than 10000 training samples are available.

\begin{table}[ht]
    \centering
    \resizebox{0.6\textwidth}{!}{%
        \begin{tabular}{c l c ccc c ccc c ccc}
            \toprule
                 & $N$ & Pointflow & \multicolumn{3}{c}{Alexnet} & & \multicolumn{3}{c}{VGG} & & \multicolumn{3}{c}{Resnet}\\ 
                  \cmidrule{4-6} \cmidrule{8-10} \cmidrule{12-14} 
                 & & & TL & FT & SC & & TL & FT & SC & & TL & FT & SC  \\ 
                 \midrule
                $(r,c)$ & 0 & 
                 \bf{0.66} &
                 & & & &
                 & & & &
                 & & \\
                 & 10 & 
                 &
                 826  & 524  & 1748 & &
                 1581 & 754  & 1748 & &
                 5256 & 3358 & 2590 \\
                 & 100 & 
                 &
                 314  & 66   & 291 & &
                 448  & 69   & 215 & &
                 2064 & 1124 & 793 \\
                 & 1000 & 
                 &
                 183  & 23  & 9.0 & &
                 189  & 6.5 & 7.6 & &
                 1092 & 48  & 45 \\
                 & 10000 & 
                 &
                 140 & 31  & 4.8 & &
                 71  & 2.9 & 1.2 & &
                 814 & 3.7 & 4.4 \\
                 & 100000 & 
                 &
                 67  & 17   & 1.5  & &
                 40  & 0.68 & \bf{0.42} & &
                 826 & \bf{0.51} & 0.93 \\
                 \midrule
                $r$ & 0 & 
                 \bf{0.26} &
                 & & & &
                 & & & &
                 & & \\
                 & 10 & 
                 &
                 66  & 65 & 75 & &
                 75  & 69 & 75 & &
                 110 & 86 & 82 \\
                 & 100 & 
                 &
                 36 & 5.9 & 42 & &
                 52 & 5.6 & 40 & &
                 42 & 25  & 31 \\
                 & 1000 & 
                 &
                 26 & 2.4  & 1.3  & &
                 20 & 0.57 & 0.62 & &
                 24 & 2.2  & 1.3 \\
                 & 10000 & 
                 &
                 16  & 2.0  & 1.5  & &
                 9.9 & \bf{0.23} & \bf{0.11} & &
                 19  & \bf{0.19} & 0.27\\
                 & 100000 & 
                 &
                 6.8 & 1.7   & 0.69  & &
                 4.9 & \bf{0.075} & \bf{0.051} & &
                 19  & \bf{0.066} & \bf{0.078} \\
                \midrule
                $c$ & 0 & 
                 \bf{0.40} &
                 & & & &
                 & & & &
                 & & \\
                 & 10 & 
                 &
                 759  & 454  & 1673 & &
                 1506 & 675  & 1673 & &
                 5142 & 3206 & 2507 \\
                 & 100 & 
                 &
                 277  & 60   & 249 & &
                 393  & 63   & 175 & &
                 2020 & 1099 & 762 \\
                 & 1000 & 
                 &
                 157  & 21  & 7.7 & &
                 169  & 5.9 & 6.9 & &
                 1067 & 46  & 43 \\
                 & 10000 & 
                 &
                 118 & 29  & 3.0 & &
                 61  & 2.6 & 1.1 & &
                 795 & 3.5 & 4.1 \\
                 & 100000 & 
                 &
                 57  & 6.4  & 0.79 & &
                 36  & 0.57 & \bf{0.37} & &
                 807 & 0.45 & 0.86 \\
            \bottomrule
        \end{tabular}
    }
    \vspace{10pt}
    \caption{MSE scores of the networks compared to Pointflow on test data, computed on both $r$ and $c$ (top), just $r$ (middle), or just $c$ (bottom). Networks were trained on joint prediction of $r$ and $c$. For the separate $r$ (resp. $c$) scores, the selected networks were those providing the best $r$ (resp. $c$) error on validation data.
    }
    \label{tab: circles mse test r 1}
\end{table}

\section{Conclusion}

We analysed the amount of data required by neural networks, either shallow custom ones or deep famous ones, trained from scratch, finetuned, or transfer-learned, to compete with optimal or state-of-the-art traditional data-agnostic methods based on mathematical data generation models. To do so, we mathematically generated data, and fed various amounts of samples to the networks for training. We found that tens of thousands of data examples are needed for the networks to be on par or beat the traditional methods, if they are able to. For mathematical accuracy, we did not investigate real-world problems, which are commonly harder with less accurate or non-existent mathematical data generation models, but more data should be needed in those complex tasks. We have empirically derived a simple criterion, enabling researchers working on tasks where data is not easily available, to choose whether to use model-based traditional methods, by using either preexisting or newly created data generation models, or simply feed data to deep neural networks.

\appendix

\section{Pointflow implementation details}
\label{sec: appendix pointflow details}

The pointflow dynamics are implemented by discretising time and approximating the time derivative with a forward finite difference scheme, although it could be improved with a Runge-Kutta 4 implementation \cite{butcher2008numerical}. Given the small magnitudes of the fields, we found that a large time step $dt = 50$ works well. We define three thresholds, $\tau_l = 0.9$ for $C_l$, $\tau_s = 10^{-6}$ for $C_s$, and $\tau_{len}=0.001$. we consider having looped $C_l$ if a point reaches a previous point within squared Euclidean distance $\tau_l$ while having on the trajectory between the looping points at least one point with squared distance to them of at least $\tau_l$. A trajectory is stuck if it reaches a point where the current flow $V$ has small magnitude $\lVert V\rVert_2^2\le \tau_s$. Each flow is run for $N_i = 1000$ iterations, and trajectories shorter than $\tau_{len}$ are discarded, e.g. trajectories of type $C_s$. We used $\sigma_{Pf} = 5$ for blurring out the noise before computing the fields. The implemented pointflow algorithm for finding contours in our circle images is presented in Algorithm \ref{alg: pointflow contour}.

\begin{algorithm}
    \caption{Contour integration with Pointflow on image $I$}\label{alg: pointflow contour}
    \begin{algorithmic}
        \State Compute $I_b = g_{\sigma_{Pf}} * I$
        \State Compute $V_a = \nabla\lVert\nabla I_b\rVert_2$ and $V_r = \nabla  I_b^\top$
        \State Compute $V_+ = \tfrac{1}{2}(V_a+V_r)$ and $V_- = \tfrac{1}{2}(V_a-V_r)$
        \State Choose $N_{Pf}$ random points independently and uniformly in the image domain $[-\frac{D-1}{2}, \frac{D-1}{2}]^2$
        \State Let $\mathcal{C} = []$ be an empty list of computed contours
        \For{$i=1\hdots N_{Pf}$}
            \State Let $(traj_+, C)$ be the  flow along $V_+$ starting from the $i$-th point 
            \If{$C = C_l$ and $\mathrm{length}(traj_+)\ge\tau_{len}$}
                \State Let $(traj_l, C_l)$ be the reflow along $V_+$ starting from the endpoint of $traj_+$ 
                \State $\mathcal{C}.\mathrm{append}(traj_l)$
            \ElsIf{$C = C_o$  and $\mathrm{length}(traj_+)\ge\tau_{len}$}
                \State Let $(traj_-, C_-)$ be the reflow along $V_-$ starting from the endpoint of $traj_+$
                \If{$C_- = C_l$ and $\mathrm{length}(traj_-)\ge\tau_{len}$}
                    \State Let $(traj_l, C)$ be the reflow along $V_-$ starting from the endpoing of $traj_-$
                    \State $\mathcal{C}.\mathrm{append}(traj_l)$
                \EndIf
            \EndIf
        \EndFor
        \State Return $\mathcal{C}$
    \end{algorithmic}
\end{algorithm}

After computing the list of contours $\mathcal{C}$ in the image $I$, we estimate the radius using the average curve length $\hat{r} = \tfrac{1}{2\pi}\sum_{i=1}^{|\mathcal{C}|} \mathrm{length}(\mathcal{C}_i)$.  Since the average of the points did not yield the best estimation of the circle centre, we estimate it instead using least squares. The equation of a circle is naturally given by $(x-c_x)^2 + (y-c_y)^2 = r^2$, which can be written as $\theta_1x + \theta_2y + \theta_3 = x^2 + y^2$, where $\theta_1 = 2c_x$, $\theta_2 = 2c_y$, and $\theta_3 = r^2 - c_x^2 - c_y^2$. We can thus estimate for each contour $\theta = (\theta_1, \theta_2, \theta_3)^\top$ by least squares as $\hat{\theta} = A^\top (AA^\top)^{-1}B$, with $A_{i,:} = (x_i, y_i, 1)$ and $B_i = x_i^2 + y_i^2$ and $i$ ranging in the number of computed points on the contour. From $\hat{\theta}$ we can estimate $\hat{c} = (\tfrac{\theta_1}{2}, \tfrac{\theta_2}{2})$. The final centre estimation is then given by the average of this estimation over all contours. Note that we can also estimate $r$ using $\theta_3$ but we found that it did not outperform the lenght strategy so we do not use it.

\subsubsection{Acknowledgements} This work is in part supported by the French government under management of Agence Nationale de la Recherche as part of the "Investissements d'avenir" program, reference ANR-19-P3IA-0001 (PRAIRIE 3IA Institute).

%
%
%
\bibliographystyle{splncs04}
\bibliography{references}
%





\end{document}